\documentclass{article}

\PassOptionsToPackage{numbers,compress}{natbib}
\usepackage[main, final]{neurips_2026}
\usepackage{environ}

\usepackage[utf8]{inputenc} 
\usepackage[T1]{fontenc}    
\usepackage{hyperref}       
\usepackage{url}            
\usepackage{booktabs}       
\usepackage{array}          
\usepackage{amsfonts}       
\usepackage{nicefrac}       
\usepackage{microtype}      
\usepackage{xcolor}         
\definecolor{lightyellow}{rgb}{1.0,1.0,0.85}
\usepackage{graphicx}       
\usepackage{float}          
\usepackage{wrapfig}        
\usepackage[most]{tcolorbox}

\usepackage{makecell}
\usepackage{array}
\usepackage{tabularx}
\usepackage{adjustbox}
\usepackage{multirow}
\usepackage{multicol}
\usepackage{xurl}
\usepackage{colortbl}

\title{From Failure Taxonomy to Intervention: \\A Diagnostic Methodology for Industry-Scale AVLM in Video and Live-Streaming Platform Moderation}

\author{
Shuchang Ye$^{1,2}$ \quad Jinqiang Yu$^{1}$ \quad \textbf{Zhujun Xiao}$^{1}$ \quad \textbf{Yajing Kong}$^{1}$ \\ \textbf{Yist Y. Lin}$^{1}$  \quad \textbf{Yang Ma}$^{1}$ \quad \textbf{Jiaxi Liu}$^{1}$ \quad \textbf{Xiaolei Xu}$^{1}$ \quad \textbf{Zheng Yu}$^{1}$\vspace{1em}
\\$^1$TikTok   $^2$The University of Sydney
}

\begin{document}

\maketitle
\raggedbottom

\begin{abstract}
Industry-scale video and live-streaming moderation imposes requirements that are difficult to satisfy with generic pretrained public models or external APIs, including adaptation to platform-specific data distributions, policy-specific objectives, and product-level safety constraints. As a result, platforms must undertake internal model development, naturally turning to shared public research for guidance. However, existing multimodal foundation-model studies primarily report architectures, training recipes, data scaling strategies, and benchmark results, but provide less systematic guidance on how failures should be localized and translated into targeted model-development interventions. Interventions are essential because deployment failures are rarely self-explanatory. Similar failures can originate from different causes. Without targeted interventions, improvement reduces to heuristic trial-and-error, where benchmark improvements are weakly attributable, and failures are difficult to trace to their underlying causes. To address this gap, we present a diagnostic methodology for industry-scale Audio-Visual-Language Models~(AVLM) development. The methodology maps model failures into a taxonomy of observable failure signatures and links each class of failure to an intervention space. We instantiate this methodology through the development and alignment of the first AVLM foundation model trained on trillion-scale data for general content understanding and downstream moderation on video and live-streaming platforms. The resulting system is deployed in production across more than 100 regions and more than 100 languages, on a platform serving billions of users and supporting tens of billions of quarterly video uploads and live streaming. 
\end{abstract}

\section{Introduction}

Modern video and live-streaming platforms host massive volumes of user-produced multimodal content, making automated moderation a core infrastructure requirement for maintaining community safety at scale~\cite{aldahoul2024advancing,huang2025content,palla2025policy}. The scale of this requirement is evident from industry transparency disclosures. Earlier reports emphasized the human labor burden of moderation, describing workforces of approximately $15{,}000$ reviewers~\cite{facebook_moderation}. Whereas more recent disclosures highlight the degree of operational automation, reporting roughly $112$ million pieces of violating content removed within six months, with $93.8\%$ actioned without human review~\cite{tiktok_moderation}. In policy-sensitive moderation, externally hosted APIs and generic pretrained models provide limited control over the operational and behavioral factors that determine deployment quality, including latency, cost, failure modes, and policy alignment. This motivates the development of platform-controlled  Audio-Visual-Language Models~(AVLMs) that can be optimized for reliability, scalability, and alignment with evolving moderation policies.

\begin{figure}[t]
  \centering
  \includegraphics[width=0.95\linewidth,height=0.55\textheight,keepaspectratio]{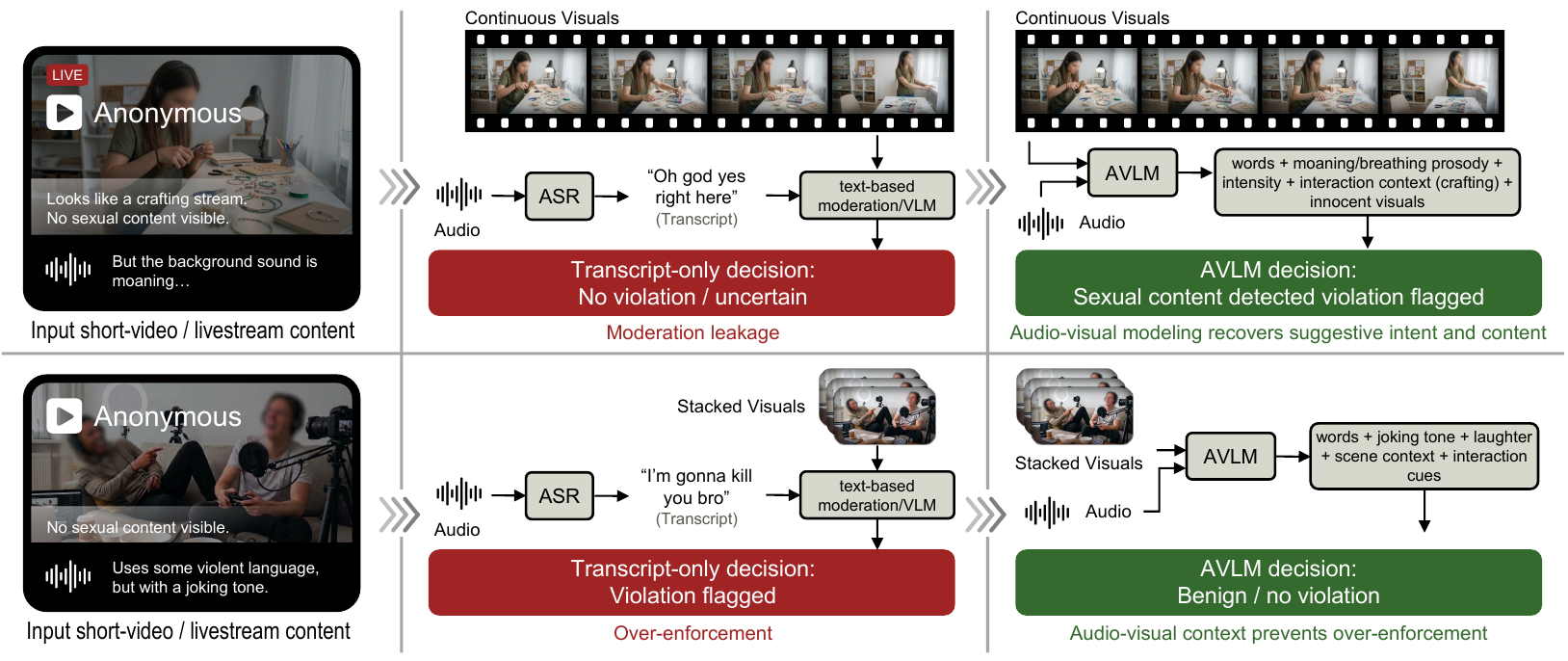}
  \caption{Audio evidence beyond transcription in video and live-streaming moderation. The upper example illustrates moderation leakage when policy-relevant acoustic evidence is lost during transcription. The lower example illustrates overkill when risky words are preserved in transcript form but the original audio-visual context indicates benign intent. (\textit{Privacy is protected through generative face anonymization.})}
  \label{fig:background}
\end{figure}

Existing AI-assisted moderation systems for video and live-streaming platforms commonly adopt Vision-Language Model~(VLM)-based pipelines~\cite{mllm_moderation}, where visual frames serve as the primary perceptual input for policy reasoning. In such pipelines, audio is often represented indirectly through ASR transcripts, which convert speech into textual evidence for the moderation model. This representation captures violations expressed through recognized words, but neglects acoustic and cross-modal evidence such as prosody, vocal intensity, non-speech sounds, environmental audio, and correspondences between sound and visual context. Figure~\ref{fig:background} illustrates two failure modes: Transcript-based systems may under-enforce when policy-relevant evidence is present in the audio signal but absent from the transcript, or over-enforce when risky words are disambiguated by the original audio-visual context.

These limitations motivate a shift from transcript-based audio modeling to Audio-Visual-Language Models (AVLMs) that jointly integrate speech, non-speech acoustic cues, visual context, and contextual metadata for moderation. However, existing literature in multimodal foundation models primarily focuses on model-building procedures, emphasizing architecture design, modality alignment, data scaling, staged training, and benchmark performance. Representative works have established visual instruction tuning for VLM alignment~\cite{llava}, scalable transfer across image and video scenarios~\cite{llava-onevision}, multimodal instruction tuning over video and audio~\cite{videollama2}, and general-purpose omni-modal modeling~\cite{qwen3-omni}. A central deployment question remains rarely discussed: \textit{How can model failures be diagnosed in ways that identify appropriate interventions?}

This question is important in industry internal model development because downstream policy errors often fail to reveal their underlying cause. A single moderation failure could stem from a variety of distinct technical issues. Without a systematic diagnostic framework, performance improvement can be reduced to heuristic trial-and-error, where benchmark improvements are weakly attributable, and failures are difficult to trace to their underlying causes.

To address this gap, we introduce a systematic diagnostic methodology for industry-scale AVLM development, in which observed failures are categorized into interpretable failure taxonomies and translated into actionable development interventions. We instantiate this methodology across the development and alignment lifecycle of an AVLM foundation model for a large-scale video and live-streaming platform. The resulting system supports over 100 regions and is designed for noisy, ambiguous, and highly diverse content drawn from global platform traffic. 

\section{AVLM Development for Industry-Scale Video and Live-Streaming Platforms}

\begin{figure}[t]
  \centering
  \includegraphics[width=0.95\linewidth,height=0.55\textheight,keepaspectratio]{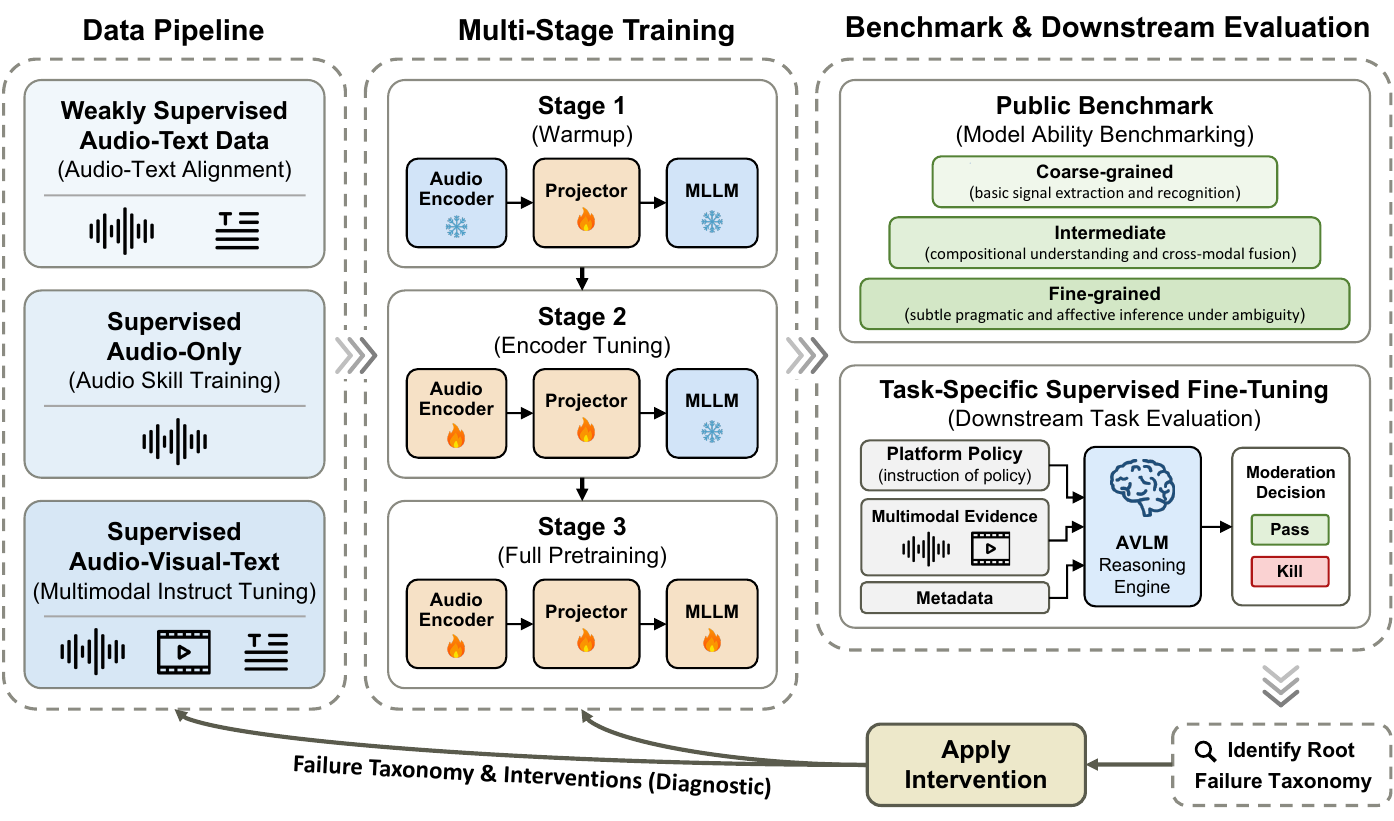}
  \caption{Overview of the proposed AVLM development and diagnostic framework. The framework couples a staged data pipeline, multi-stage training, and benchmark\&downstream evaluation for industrial AVLM development. It also exposes a diagnostic feedback loop, in which observed evaluation failures are abstracted into root failure taxonomies and mapped to interventions.}
  \label{fig:avlm}
\end{figure}

In this section, we present an overview of the development pipeline of AVLM, as  shown in Figure~\ref{fig:avlm}.

\subsection{Supervision Regimes for Platform Audio}

Training data is categorized into three supervision regimes that progressively shift from scalable audio acquisition to task-oriented multimodal reasoning, as shown in Table~\ref{tab:supervision_regimes}.

\paragraph{Weakly supervised audio-text data.}
The first regime contains approximately \textbf{1 million hours} of pseudo-labeled audio-text pairs from online ASR sources. It provides broad multilingual speech coverage for initial audio capability acquisition. We apply automatic quality control, including agreement-based filtering between audio language identification and ASR outputs, and entropy-based filtering to remove repetitive or low-information transcripts.

\paragraph{Supervised audio-only data.}
The second regime contains approximately \textbf{200,000 hours} of supervised audio data from in-house video and live scenarios. It moves training beyond transcript reconstruction toward broader auditory understanding, covering ASR, AST, diarization-ASR, speech summarization, intent understanding, harmful-content identification, and audio event summarization. Compared with the weakly supervised corpus, this regime is smaller but provides denser supervision for speaker behavior, paralinguistic cues, and mixed speech/non-speech acoustic patterns.

\paragraph{Supervised audio-visual instruction data.}
The third regime introduces audio-instruction following and cross-modal collaboration, aiming to integrate audio-visual reasoning while preserving established visual-textual performance.
Transcript pairs broaden lexical exposure, whereas event and intent targets make acoustic properties useful beyond transcription. Audio-visual targets additionally require interpreting whether the modalities reinforce or disambiguate each other.

\begin{table}[htbp]
\centering
\small
\caption{Training supervision regimes used in the AVLM development pipeline.}
\label{tab:supervision_regimes}
\setlength{\tabcolsep}{5pt}
\begin{tabular}{lll}
\toprule
\textbf{Regime} & \textbf{Primary tasks} \\
\midrule
T0: weak audio-text data & ASR-style pseudo-labeling  \\
T1: supervised audio data & ASR, audio event perception, diarization, AST  \\
T2: supervised audio-visual data & AV grounding, VQA/AVQA, instruction following \\
\bottomrule
\end{tabular}
\end{table}


\subsection{Architecture}

The AVLM couples an in-house audio encoder with a pretrained language or vision-language backbone via a lightweight projector. Raw waveform segments are encoded into audio representations, mapped into the backbone embedding space, and jointly processed with text tokens and, visual tokens from sampled frames. This formulation enables speech content, non-speech acoustic cues, visual context, and cross-modal interactions to be modeled within a shared multimodal language-model decoder.
The encoder must retain relevant acoustic cues, the projector must make them accessible, and the decoder must use them. Reliable lexical recovery alone does not test whether this chain supports other policy-relevant evidence.

\subsection{Three-stage AVLM Training}

\paragraph{Stage I. Weakly supervised audio text alignment.}
The first stage establishes the audio language alignment using large scale weakly supervised audio text data. Its objective is to map audio encoder representations into the pretrained language backbone space and acquire broad audio text correspondence from high coverage pseudo supervision.

\paragraph{Stage II. Audio perception and reasoning adaptation.}
The second stage adapts the audio encoder and projection modules with supervised audio data. This stage strengthens speech understanding, non-speech perception, and audio reasoning beyond transcript reconstruction. It also improves sensitivity to paralinguistic cues, speaker behavior, and environmental sounds, thereby reducing over-reliance on ASR-style speech supervision.

\paragraph{Stage III. Joint multimodal capability consolidation.}
The third stage trains the full model using mixed supervision spanning audio, vision, text, and audio-visual instructions. This stage aims to preserve the vision-language capabilities inherited from the VLM backbone while further improving audio understanding and audio-visual reasoning.
Full-model updates change the decoder interface against which audio representations were learned. Each transition therefore requires checking both newly acquired capability and regression on earlier tasks.

Foundation-model pretraining uses one epoch, a learning rate of $10^{-5}$, and global batch size 1,024. Moderation fine-tuning uses two epochs, a learning rate of $5\times10^{-6}$, per-device batch size 2, and eight gradient-accumulation steps.

\section{Experiment Setup}
\label{sec:experiment_setup}

Our experiments evaluate an in-house AVLM designed for audio perception, audio-visual reasoning, and platform-oriented multimodal understanding. We evaluate the model in two dimensions: general multimodal understanding and downstream moderation performance. 

\subsection{Evaluation Datasets}

For downstream moderation evaluation, we use anonymized in-house video and live-streaming benchmarks covering two primary policy families, denoted as \textit{Anon.S} representing sexual-related violation and \textit{Anon.V} representing violence-related violation. These benchmarks evaluate policy-aligned audio-visual AI moderation performance under large-scale complex-context real-world scenarios.

For public audio-visual evaluation, we use AVHBench~\cite{avhbench} and OmniBench~\cite{omnibench}. These benchmarks evaluate zero-shot audio-visual understanding, audio-visual matching, omni-modal reasoning, and instruction-following behavior.

For audio-only evaluation, we use FLEURS~\cite{fleurs}, MMAU~\cite{mmau}, and MMSU~\cite{mmsu}. FLEURS evaluates multilingual speech recognition across 18 languages. MMAU and MMSU evaluate broader audio perception, reasoning, and knowledge-oriented understanding.

For general multimodal evaluation, we use benchmarks covering STEM reasoning, general VQA, document and chart understanding, OCR, counting, multi-image reasoning, spatial understanding, and video reasoning. These include MMMU~\cite{mmmu}, RealWorldQA~\cite{realworldqa}, MMStar~\cite{mmstar}, GQA~\cite{gqa}, SEEDBench~\cite{seed-bench}, POPE~\cite{pope}, ScienceQA~\cite{scienceqa}, MME~\cite{mme}, AI2D~\cite{ai2d}, OCRBench~\cite{ocrbench}, CountBench~\cite{countbench}, DocVQA~\cite{docvqa}, ChartQA~\cite{chartqa}, TextVQA~\cite{textvqa}, InfoVQA~\cite{infovqa}, BLINK~\cite{blink}, MUIRBENCH~\cite{muirbench}, ERQA~\cite{erqa}, EmbSpatialBench~\cite{embspatialbench}, MVBench~\cite{mvbench}, VideoMME~\cite{videomme}, and LVBench~\cite{lvbench}.

\subsection{Metrics}

For downstream moderation benchmarks, we report PR-AUC because the evaluated policy behaviors are class-imbalanced. For FLEURS, we report WER/CER, where lower values indicate better multilingual speech recognition. For MMAU, MMSU, AVHBench, OmniBench, and most general multimodal benchmarks, we report the benchmark-defined accuracy or score. For document-oriented benchmarks, we use the benchmark-standard metrics, including ANLS and exact match where applicable.


\section{Failure Taxonomy and Interventions}

This section presents representative failure taxonomies observed during AVLM development and maps each failure type to targeted interventions within the development lifecycle. 
These categories can overlap. Auditing retained support before testing further exposure separates missing examples from under-training. Re-evaluation after instruction tuning then distinguishes new regression from earlier supervision bottlenecks.

\subsection{Selection-Induced Support Erosion}

In the Stage~I pretraining data pipeline, large weakly supervised corpora are typically collected from platform traffic to provide broad coverage of real-world, user-produced content. However, this raw data is systematically reshaped by media availability, validity checks, audio-visual alignment constraints, and quality-control filters. Because these selection mechanisms are not distribution-neutral, they preferentially retain clean, high-resource, and easily aligned examples. Consequently, they actively suppress the noisy, long-tail cases that are critical for learning nuanced moderation boundaries.

Let $p_{\mathrm{traffic}}(x)$ denote the underlying distribution of raw platform content. The data pipeline subjects each sample $x$ to a sequence of filters, which we can represent as an aggregate retention probability $S_{\mathrm{retain}}(x)$. The final, post-filter training distribution $q(x)$ is therefore:
\[
q(x) = \frac{p_{\mathrm{traffic}}(x) S_{\mathrm{retain}}(x)}{\mathbb{E}_{x \sim p_{\mathrm{traffic}}}[S_{\mathrm{retain}}(x)]}
\]
When $S_{\mathrm{retain}}(x)$ is lower for minority languages, poor audio conditions, or complex multimodal edge cases, raw corpus size can overstate the coverage available for learning. Therefore, diagnostics and interventions must target the post-filter training distribution $q(x)$, rather than $p_{\mathrm{traffic}}(x)$. Table~\ref{tab:data_controls} summarizes the diagnostic signals of this erosion and their corresponding controls.
We compare pre- and post-filter distributions by region, language, audio quality, supervision source, and policy tails. Reweighting can increase exposure to retained cases, but cannot recover a case type with zero retained support. That requires revising collection or filtering.

\begin{table}[htbp]
\centering
\small
\caption{Diagnostic signals for selection-induced support erosion. Controls must target the post-selection training distribution rather than the raw collection corpus.}
\label{tab:data_controls}
\setlength{\tabcolsep}{4pt}
\renewcommand{\arraystretch}{1.25}
\begin{tabularx}{\linewidth}{
>{\raggedright\arraybackslash}p{0.30\linewidth}
>{\raggedright\arraybackslash}p{0.40\linewidth}
>{\raggedright\arraybackslash}X}
\toprule
\textbf{Diagnostic Signal} & \textbf{Underlying Mechanism} & \textbf{Corrective Control} \\
\midrule
Highly skewed region and language histograms
& Filtering over-represents high-resource demographics, degrading minority-language and region-specific acoustic performance.
& Post-filter region and language balancing. \\
Raw traffic is dominated by benign "head" content
& Abundant supervision on frequent, low-risk patterns crowds out exposure to rare, high-risk policy violations.
& Content-aware sampling and upweighting of tail events. \\
Disproportionate removal of weakly aligned media
& Strict quality control systematically deletes the noisy, ambiguous samples that models need for robust real-world deployment.
& Alignment-aware filter calibration and post-filter auditing. \\
\bottomrule
\end{tabularx}
\end{table}

\begin{tcolorbox}[colback=lightyellow,colframe=black,arc=4pt,boxsep=1pt]
\paragraph{\textbf{\textit{Failure Taxonomy} 1. }}
Data curation pipelines act as non-neutral filters, transforming broad raw traffic into a narrow training distribution. This strips away the critical long-tail and noisy examples required for robust moderation capability.

\paragraph{\textbf{\textit{Intervention}.}}
Treat the data pipeline as a biased sampling mechanism. Reconstruct the necessary distribution through region-aware sampling, content-driven upweighting, and calibrating quality filters to retain policy-relevant edge cases.
\end{tcolorbox}

\begin{figure}[b]
  \centering
  \includegraphics[width=0.95\linewidth,height=0.55\textheight,keepaspectratio]{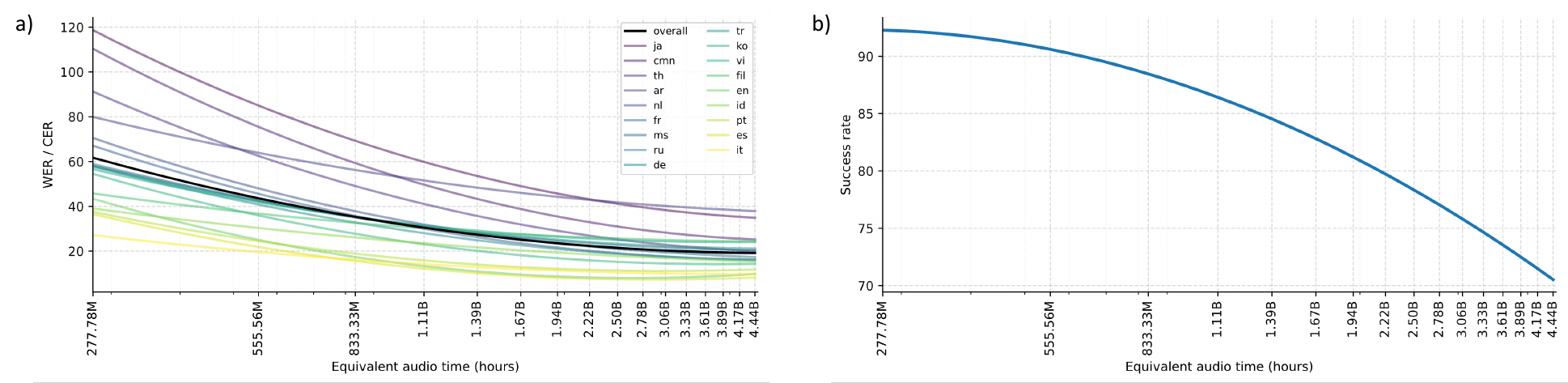}
  \caption{Effect of scaling equivalent audio training time. a) FLEURS WER/CER across common languages and overall. b) MMSU question-answering success rate.}
  \label{fig:scaling}
\end{figure}

\subsection{Regime-Limited Audio-Language Alignment}

Weak audio-text pretraining in Stage~I serves as the first diagnostic point in AVLM development because it determines whether acoustic representations can be reliably interfaced with the language backbone. The objective of this stage is not limited to transcription accuracy. It is to establish an audio-language mapping that later stages can inherit for reasoning, instruction following, and multimodal grounding. When this pretraining is not converged, failures observed in later supervised stages may originate from unstable cross-modal representations rather than from insufficient reasoning supervision.

Figure~\ref{fig:scaling} illustrates this transition through the scaling behavior of weak audio-text pretraining. Continued held-out improvement supports extending weak pretraining. A stable plateau motivates testing higher-information supervision, but does not by itself identify a unique bottleneck. Strong transcription with persistently weak audio question answering (AQA) after this plateau distinguishes the next diagnosis from insufficient Stage-I exposure.

\begin{tcolorbox}[colback=lightyellow,colframe=black,arc=4pt,boxsep=1pt]
\paragraph{\textbf{\textit{Failure Taxonomy} 2. }}
The audio-language mapping remains unreliable when weak pretraining has not reached the exposure regime required for stable cross-modal alignment.

\paragraph{\textbf{\textit{Intervention}.}}
Use scaling curves as transition tests: scale weak pretraining while alignment error decreases, then replace additional exposure with higher-information reasoning and multimodal supervision.
\end{tcolorbox}

\subsection{Proxy-Sufficient Representation Collapse}

After Stage~I weak audio-text pretraining is saturated, the next diagnostic question is whether the learned audio representation contains information that is usable by the decoder beyond lexical recovery. Stage~I audio-to-text pretraining rewards representations that preserve lexical content, but may underweight semantic intent, paralinguistic cues, non-speech events, background context, and audio-visual consistency. These non-speech signals are often necessary for Stage~II audio reasoning and downstream moderation.

This mismatch can make the audio pathway overfit to speech recognition. The model is optimized for recognizing lexical content, as reflected by lower WER or CER, but the resulting audio features may not preserve sufficient audio information for downstream reasoning, instruction following, or cross-modal grounding. As a result, ASR performance can improve while AQA accuracy degrades.

Cross-Modal Continuation~\footnote{Cross-modal continuation refers to generating or extending content in one modality conditioned on context from another modality. In our case, the model generates text conditioned on audio.} (CMC) addresses this representation-level mismatch before Stage~III instruction tuning by requiring the decoder to continue language from the audio context. The objective does not directly optimize downstream policy decisions. Instead, it aligns projected audio representations with the language-generation interface of the backbone, with the aim of helping the audio pathway use the instruction-following and reasoning capabilities of the backbone. Table~\ref{tab:cmc_effect} shows that CMC improves AQA performance while preserving or strengthening transcription quality.
CMC uses timestamp-aligned pairs split at a valid word boundary, with an audio prefix as input and transcript suffix as target. Low-confidence alignments and too-short examples are removed. CMC comprises 20\% of the audio-data mixture. Joint ASR and AQA improvements show no inherent conflict, but do not isolate representation changes from data selection or continuation supervision.

\begin{table}[htbp]
\centering
\small
\caption{Effect of cross-modal continuation (CMC) on transcription and AQA-style capability.}
\label{tab:cmc_effect}
\begin{tabular}{lcccc}
\toprule
\textbf{Model} & \textbf{LibriSpeech-Other $\downarrow$} & \textbf{FLEURS overall $\downarrow$} & \textbf{MMAU $\uparrow$} & \textbf{MMSU $\uparrow$} \\
\midrule
Without CMC & 13.40 & 29.26 & 34.2 & 18.98 \\
With CMC    & 10.41 & 17.98 & 52.5 & 40.98 \\
\bottomrule
\end{tabular}
\end{table}

\begin{tcolorbox}[colback=lightyellow,colframe=black,arc=4pt,boxsep=1pt]
\paragraph{\textbf{\textit{Failure Taxonomy} 3. }}
A representation can become sufficient for a dominant proxy objective while under-encoding information needed for reasoning, instruction following, and decision use.

\paragraph{\textbf{\textit{Intervention}.}}
Add cross-modal generative supervision that trains the modality representation to support decoder continuation beyond proxy recovery.
\end{tcolorbox}

\subsection{Instruction Tuning Can Trade Perception for Reasoning}

Stage~III task-oriented training emphasizes instruction-following and multimodal reasoning. A typical failure mode is that reasoning and instruction-following abilities improve while perception is weakened. Because the model is optimized by response-level supervision, where correct answers may be obtained from higher-level semantic cues, visual context, metadata, or dataset priors without requiring full retention of speech-perception fidelity.

Table~\ref{tab:stage3_tradeoff} shows that instruction tuning improves AQA reasoning but can degrade ASR performance. CMC partially mitigates this degradation, consistent with aligning the intermediate audio representation with the language-model decoding space before Task-oriented training. Since CMC trains the model to perform language continuation from audio context, it encourages the audio pathway to preserve language-predictive lexical, semantic, and acoustic evidence. This may provide a stronger initialization for instruction tuning and reduce the risk that Stage~III optimization relies on shortcuts that bypass detailed audio perception.
FLEURS error remains 33.87 with CMC after Stage~III, versus 29.26 in the Stage-II baseline and 17.98 at the pre-instruction CMC checkpoint in Table~\ref{tab:cmc_effect}. Thus, mitigation is partial, and improved reasoning does not imply retention of the strongest earlier speech capability.

\begin{table}[htbp]
\centering
\small
\caption{Instruction tuning improves AQA reasoning while weakening speech perception.}
\label{tab:stage3_tradeoff}
\begin{tabular}{lcccc}
\toprule
\textbf{Checkpoint} & \textbf{LibriSpeech-Other $\downarrow$} & \textbf{FLEURS overall $\downarrow$} & \textbf{MMAU $\uparrow$} & \textbf{MMSU $\uparrow$} \\
\midrule
Stage~II & 13.40 & 29.26 & 34.2 & 18.98 \\
Stage~III without CMC & 13.66 & 34.90 & 71.0 & 51.06 \\
Stage~III with CMC    & 12.89 & 33.87 & 72.4 & 51.34 \\
\bottomrule
\end{tabular}
\end{table}

\begin{tcolorbox}[colback=lightyellow,colframe=black,arc=4pt,boxsep=1pt]
\paragraph{\textbf{\textit{Failure Taxonomy} 4.}} Instruction tuning can trade perception for instruction following and reasoning. High-quality multimodal instruction data improves instruction-following and reasoning while weakening perception.

\paragraph{\textbf{\textit{Intervention}.}} Apply cross-modal continuation before Stage~III to strengthen decoder-aligned audio conditioning. 
\end{tcolorbox}

\subsection{The Perception Assumption: Does Better ASR Guarantee Better Moderation?}

Across Stage~II audio adaptation and Stage~III multimodal consolidation, an intuitive hypothesis is that improving fundamental audio perception, essentially making the model a better transcriber, will result in better downstream AI moderation. This assumption is inherited from legacy safety pipelines, where audio moderation is typically handled by cascading an ASR module to convert audio signals into textual transcripts before applying a VLM or LLM.

However, downstream AI moderation relies on reasoning about semantic intent, paralinguistic evidence, acoustic context, audio-visual grounding, and policy-specific decision boundaries. Consequently, a model can significantly improve its generic perception capabilities while actively degrading its downstream moderation performance, as illustrated in Table~\ref{tab:asr_vs_downstream}.
The architecture, initialization, optimization, and total budget are fixed while ASR, non-ASR audio, and AQA shares change. The result therefore identifies a recipe-level trade-off. WER/CER weights lexical errors without measuring their policy consequences, so it cannot alone determine the best moderation recipe.

\begin{table}[htbp]
\centering
\small
\caption{Recipe-level trade-off under fixed training budget. The perception-heavy recipe emphasizes lexical recovery. The reasoning-balanced recipe upweights non-ASR audio tasks and AQA. LS denotes LibriSpeech.}
\label{tab:asr_vs_downstream}
\setlength{\tabcolsep}{3.5pt}
\begin{adjustbox}{max width=\linewidth}
\begin{tabular}{lccccccc}
\toprule
\multirow{2}{*}{\textbf{Recipe}}
& \multicolumn{3}{c}{\textbf{ASR} $\downarrow$}
& \multicolumn{2}{c}{\textbf{AQA} $\uparrow$}
& \multicolumn{2}{c}{\textbf{Moderation} $\uparrow$} \\
\cmidrule(lr){2-4} \cmidrule(lr){5-6} \cmidrule(lr){7-8}
& \textbf{LS-Clean} & \textbf{LS-Other} & \textbf{FLEURS}
& \textbf{MMAU} & \textbf{MMSU} & \textbf{Anon.S} & \textbf{Anon.V} \\
\midrule
Perception-heavy & 7.21 & 12.85 & 20.29 & 43.0 & 40.0 & 0.459 & 0.517 \\
Reasoning-balanced & 7.91 & 12.89 & 29.90 & 72.5 & 51.34 & 0.565 & 0.580 \\
\bottomrule
\end{tabular}
\end{adjustbox}
\end{table}

\begin{tcolorbox}[colback=lightyellow,colframe=black,arc=4pt,boxsep=1pt]
\paragraph{\textbf{\textit{Failure Taxonomy} 5. }}
A model can improve generic perception ability while degrading downstream AI moderation, although this comparison does not establish competition within the learned representation.

\paragraph{\textbf{\textit{Intervention}.}}
Frame model selection as a multi-objective optimization problem. Target the Pareto frontier between generic perception and complex reasoning, requiring minimum capability scores before selecting for downstream moderation performance.
\end{tcolorbox}



\definecolor{oursblue}{RGB}{214,239,246}
\newcommand{\ourscell}[1]{\cellcolor{oursblue}#1}
\newcommand{\best}[1]{\textbf{#1}}

\begin{table}[htbp]
\centering
\caption{Overall multimodal evaluation on public benchmarks. Benchmarks are grouped by main modality category and visual subcategory. Bold indicates the best result among the three models.}
\label{tab:overall_mm_eval}
\vspace{2mm}
\small
\setlength{\tabcolsep}{3.5pt}
\begin{adjustbox}{max width=\textwidth}
\begin{tabular}{llllccc}
\toprule
Category & Subcategory & Benchmark & Metric 
& \ourscell{Our AVLM-2B} 
& Qwen2.5-Omni-3B 
& Gemma-4-E2B-it \\
\midrule

\multirow{2}{*}{Audio-Visual}
& \multirow{2}{*}{AV Understanding}
& AVHBench & Accuracy
& \ourscell{61.60} & \best{71.00} & 50.72 \\
& 
& OmniBench & Accuracy
& \ourscell{43.86} & \best{43.96} & 21.71 \\

\midrule
\multirow{3}{*}{Audio}
& Speech Recognition
& FLEURS & WER/CER $\downarrow$
& \ourscell{\best{13.45}} & 41.15 & 16.03 \\
& Audio Understanding
& MMAU & Accuracy
& \ourscell{68.30} & \best{72.80} & 55.90 \\
& Speech Understanding
& MMSU & Accuracy
& \ourscell{51.90} & \best{61.40} & 52.80 \\

\midrule
\multirow{22}{*}{Visual}
& STEM
& MMMU & Accuracy
& \ourscell{42.67} & \best{44.56} & 34.78 \\

\cmidrule(lr){2-7}
& \multirow{7}{*}{General VQA}
& RealWorldQA & Accuracy
& \ourscell{\best{62.75}} & 58.69 & 43.92 \\
&
& MMStar & Avg. accuracy
& \ourscell{\best{57.10}} & 53.82 & 28.85 \\
&
& GQA & Accuracy
& \ourscell{\best{58.40}} & 26.46 & 39.51 \\
&
& SEEDBench & Accuracy
& \ourscell{75.10} & \best{76.17} & 64.71 \\
&
& POPE & Accuracy / F1
& \ourscell{\best{90.29 / 90.20}} & 88.23 / 87.93 & 84.01 / 84.07 \\
&
& ScienceQA & Accuracy
& \ourscell{\best{85.03}} & 81.66 & 51.21 \\
&
& MME & Score (perc. + cog.)
& \ourscell{1562 + 493} & \best{1627 + 560} & 938 + 227 \\

\cmidrule(lr){2-7}
& \multirow{7}{*}{Text/Chart/Doc}
& AI2D & Accuracy
& \ourscell{74.00} & \best{76.98} & 24.84 \\
&
& OCRBench & Score
& \ourscell{75.90} & \best{78.20} & 72.80 \\
&
& CountBench & Accuracy
& \ourscell{\best{86.97}} & 74.13 & 52.14 \\
&
& DocVQA & ANLS
& \ourscell{\best{91.51}} & 38.27 & 73.49 \\
&
& ChartQA & Relaxed accuracy
& \ourscell{\best{77.72}} & 19.84 & 51.72 \\
&
& TextVQA & VQA accuracy
& \ourscell{\best{75.52}} & 20.17 & 61.19 \\
&
& InfoVQA & ANLS
& \ourscell{\best{68.18}} & 36.33 & 39.12 \\

\cmidrule(lr){2-7}
& \multirow{2}{*}{Multi-Image}
& BLINK & Accuracy
& \ourscell{\best{46.93}} & 36.39 & 43.93 \\
&
& MUIRBench & Accuracy
& \ourscell{\best{46.88}} & 35.15 & 21.58 \\

\cmidrule(lr){2-7}
& \multirow{2}{*}{Spatial}
& ERQA & Accuracy
& \ourscell{\best{39.75}} & 36.75 & 35.50 \\
&
& EmbSpatialBench & Accuracy
& \ourscell{\best{62.72}} & 58.16 & 48.32 \\

\cmidrule(lr){2-7}
& \multirow{3}{*}{Video}
& MVBench & Accuracy
& \ourscell{59.00} & \best{65.50} & 25.24 \\
&
& VideoMME w/o subtitles & Accuracy
& \ourscell{54.81} & \best{62.37} & 42.00 \\
&
& LVBench & Accuracy
& \ourscell{38.22} & \best{46.48} & 24.92 \\

\bottomrule
\end{tabular}
\end{adjustbox}
\end{table}

\section{Evaluation}
\label{sec:evaluation}

We systematically evaluate our AVLM on public benchmarks and anonymized downstream moderation tasks, covering audio-visual understanding, audio capability, and visual capability. To contextualize performance under comparable model scale, we compare against Qwen2.5-Omni-3B and Gemma-4-E2B-it, two recent open-source multimodal baselines with similar parameter sizes.

\subsection{Overall Multimodal Evaluation}

Table~\ref{tab:overall_mm_eval} reports public audio-visual, audio, and visual benchmark results against Qwen2.5-Omni-3B and Gemma-4-E2B-it. Under a comparable model-scale setting, our AVLM demonstrates strong overall competitiveness, with particularly clear advantages in multilingual speech recognition and visual-language understanding. Despite using a compact 2B-scale architecture, our model achieves the lowest FLEURS WER/CER among the three models, showing that strong speech recognition is achievable at this model scale. The model also shows strong visual capability across general VQA, text/chart/document understanding, multi-image reasoning, and spatial reasoning. Establishing retention of inherited capability requires a comparison with the unadapted backbone. While Qwen2.5-Omni-3B remains stronger on several audio-reasoning, audio-visual, and video-centric benchmarks, and Gemma-4-E2B-it shows competitive performance on selected speech-understanding tasks, the overall results show that our AVLM provides a favorable balance between model size, audio capability, and multimodal competence.
The transcription advantage alongside weaker AQA and video scores makes interpretation of temporal multimodal evidence a separate development target. Recognition accuracy alone does not establish the reasoning needed for moderation.

\subsection{Downstream Modality Ablation}

To evaluate the impact of audio modeling on downstream moderation, we conduct a modality ablation study on anonymized platform data collected across over 100 regions. We compare a transcript-based baseline (\texttt{ASR+T}), which processes ASR-transcribed speech alongside contextual metadata, against audio modeling (\texttt{A+T}) that replaces the transcript with parsed direct real audio signals while retaining the same metadata. We evaluate \texttt{A+T} under both frozen-backbone and fully adapted configurations.
Anon.S and Anon.V contain approximately 45,500 and 117,403 samples, with 12\% and 8\% positives, from three months of traffic across over 100 languages. Each sample receives at least three independent annotations, with up to six for ambiguous cases. High-confidence positive labels must lead the next-highest label by at least two votes.

As shown in Table~\ref{tab:downstream_eval}, audio modeling yields higher observed PR-AUC compared to the transcript baseline, with full adaptation yielding the strongest performance. These gains are particularly pronounced for \texttt{Anon.S}, where policy-relevant evidence often depends on paralinguistic and non-lexical cues. 
Frozen audio improves Anon.S/Anon.V by 0.041/0.003, while full adaptation adds 0.066/0.018. Thus, input substitution alone does not account for the full improvement.

We also observe improvements on \texttt{Anon.V}, which is consistent with audio providing complementary evidence, such as tone, acoustic intensity, and background noise, beyond textualized speech. These findings are compatible with cascaded transcript-based moderation being vulnerable to both information loss from discarding critical acoustic evidence and context collapse from evaluating risky text without the necessary paralinguistic framing.
The ablation does not isolate acoustic mechanisms from transcription errors or policy adaptation. Bootstrap intervals quantify evaluation-set uncertainty, with paired intervals needed for between-setting differences.

\setlength{\textfloatsep}{12pt plus 2pt minus 2pt}
\setlength{\intextsep}{8pt plus 2pt minus 2pt}
\begin{table}[htbp]
\centering
\small
\caption{Transcript-based and audio-based moderation on the held-out policy sets. Entries show PR-AUC and aggregate 95\% bootstrap confidence intervals.}
\label{tab:downstream_eval}
\setlength{\tabcolsep}{8pt}
\begin{tabular}{lcc}
\toprule
\textbf{Setting} & \textbf{Anon.S PR-AUC $\uparrow$} & \textbf{Anon.V PR-AUC $\uparrow$} \\
\midrule
ASR+T            & 0.447 [0.402, 0.495] & 0.559 [0.520, 0.599] \\
A+T (freeze VL)  & 0.488 [0.442, 0.534] & 0.562 [0.525, 0.602] \\
A+T (full)       & 0.554 [0.509, 0.599] & 0.580 [0.542, 0.619] \\
\bottomrule
\end{tabular}
\end{table}

\subsection{Multilingual Speech Recognition}

Table~\ref{tab:fleurs_eval} reports per-language multilingual speech recognition results on FLEURS. Our AVLM obtains the best average WER/CER among the three models, with an average error rate of 13.45 compared with 41.15 for Qwen2.5-Omni-3B and 16.03 for Gemma-4-E2B-it. Across the 18 evaluated languages, our AVLM achieves the best result on 13 languages. Qwen2.5-Omni-3B performs best on English and Chinese, while Gemma-4-E2B-it performs best on Arabic, Thai, and Turkish. Overall, the results show that our audio training pipeline provides strong multilingual speech recognition performance while maintaining competitive audio-visual and visual-language capability.
Arabic error of 45.50 versus 22.05 for Gemma-4-E2B-it nevertheless motivates language-specific checks of retained training support and moderation performance. The aggregate advantage does not establish uniform reliability across languages.

\begin{table}[htbp]
\centering
\small
\caption{FLEURS WER/CER ($\downarrow$) across 18 languages. AVLM, Qwen, and Gemma denote Our AVLM-2B, Qwen2.5-Omni-3B, and Gemma-4-E2B-it. Bold indicates the best result per language.}
\label{tab:fleurs_eval}
\setlength{\tabcolsep}{3pt}
\begin{tabular}{lccc@{\hspace{10pt}}lccc@{\hspace{10pt}}lccc}
\toprule
\textbf{Lang.} & \textbf{AVLM} & \textbf{Qwen} & \textbf{Gemma} & \textbf{Lang.} & \textbf{AVLM} & \textbf{Qwen} & \textbf{Gemma} & \textbf{Lang.} & \textbf{AVLM} & \textbf{Qwen} & \textbf{Gemma} \\
\midrule
en & 6.50 & \textbf{4.18} & 9.14 & ko & \textbf{15.97} & 56.07 & 22.96 & ms & \textbf{16.98} & 36.02 & 17.49 \\
ja & \textbf{18.71} & 50.33 & 24.70 & id & \textbf{11.79} & 27.46 & 13.04 & fil & \textbf{18.30} & 67.14 & 21.87 \\
de & \textbf{8.44} & 28.90 & 13.44 & th & 10.15 & 54.97 & \textbf{9.63} & ru & \textbf{8.09} & 64.17 & 14.53 \\
es & \textbf{6.18} & 33.71 & 9.53 & fr & \textbf{9.44} & 22.58 & 14.88 & tr & 20.35 & 75.96 & \textbf{16.52} \\
ar & 45.50 & 70.30 & \textbf{22.05} & vi & \textbf{11.69} & 37.94 & 17.80 & zh & 11.25 & \textbf{5.33} & 21.50 \\
pt & \textbf{5.47} & 36.38 & 10.95 & it & \textbf{4.66} & 14.79 & 9.82 & nl & \textbf{12.70} & 54.54 & 18.76 \\
\midrule
Average & \textbf{13.45} & 41.15 & 16.03 & & & & & & & & \\
\bottomrule
\end{tabular}
\end{table}

\section{Conclusion}
\label{sec:conclusion}

This work presents a failure taxonomy for industry-scale AVLM development pipelines, categorizing recurring model failures and associating each failure class with targeted interventions. By integrating these interventions throughout the AVLM construction and alignment process, the resulting model demonstrates improved general audio-visual capability, stronger adaptation to platform-specific content, and consistent gains on downstream AI moderation tasks.

The foundation model developed in this work has since been adopted for production moderation on a global video and live-streaming platform. Sharing the development experience behind this deployment provides a practical account of the decisions and challenges involved in bringing an AVLM foundation model into industrial use.

We hope this study encourages the community to recognize model behavior intervention as an important research problem and to pursue further work in this direction.

\paragraph{Limitations.}
Due to platform privacy, safety, and policy-governance constraints, some production datasets, policy definitions, and sensitive moderation examples cannot be publicly released, and several downstream evaluations are therefore reported using anonymized policy categories.

\clearpage
\bibliographystyle{plainnat}
\bibliography{references}


\end{document}